\documentclass[sigconf,10pt]{acmart}
\usepackage{algorithm}
\usepackage{algorithmic}
\usepackage{physics}
\usepackage{amsmath}
\usepackage{tikz}
\usepackage{mathdots}
\usepackage{cancel}
\usepackage{color}
\usepackage{siunitx}
\usepackage{array}
\usepackage{multirow}
\usepackage{gensymb}
\usepackage{tabularx}
\usepackage{extarrows}
\usepackage{booktabs}
\usetikzlibrary{fadings}
\usetikzlibrary{patterns}
\usetikzlibrary{shadows.blur}
\usetikzlibrary{shapes}

\setcopyright{acmlicensed}
\copyrightyear{2025}
\acmYear{2025}
\acmDOI{XXXXXXX.XXXXXXX}

\acmConference[BIOKDD 2025]{24th International Workshop on Data Mining in Bioinformatics}{August 03,
  2025}{Toronto, Canada}

\begin{document}

\title[Deep Generative Models of Evolution]{Deep Generative Models of Evolution: SNP-level Population Adaptation by Genomic Linkage Incorporation}

\author{Julia Siekiera}
\affiliation{%
  \institution{Johannes Gutenberg University Mainz}
  \city{Mainz}
  \country{Germany}}
\email{siekiera@uni-mainz.de}

\author{Christian Schlötterer}
\affiliation{%
  \institution{Vetmeduni Vienna}
  \city{Vienna}
  \country{Austria}}
\email{christian.schloetterer@vetmeduni.ac.at}

\author{Stefan Kramer}
\affiliation{%
  \institution{Johannes Gutenberg University Mainz}
  \city{Mainz}
  \country{Germany}}
\email{kramer@informatik.uni-mainz.de}

\hypersetup{
pdfauthor={Julia Siekiera, Christian Schlötterer, Stefan Kramer},
}


\begin{abstract}
The investigation of allele frequency trajectories in populations evolving under controlled environmental pressures has become a popular approach to study evolutionary processes on the molecular level. Statistical models based on well-defined evolutionary concepts can be used to validate different hypotheses about empirical observations. Despite their popularity, classic statistical models like the Wright-Fisher model suffer from simplified assumptions such as the independence of selected loci along a chromosome and uncertainty about the parameters. Deep generative neural networks offer a powerful alternative known for the integration of multivariate dependencies and noise reduction. 
Due to their high data demands and challenging interpretability they have, so far, not been widely considered in the area of population genomics. 
To address the challenges in the area of Evolve and Resequencing experiments (E\&R) based on pooled sequencing (Pool-Seq) data, we introduce a deep generative neural network that aims to model a concept of evolution based on empirical observations over time. The proposed model estimates the distribution of allele frequency trajectories by embedding the observations from single nucleotide polymorphisms (SNPs) with information from neighboring loci. 
Evaluation on simulated E\&R experiments demonstrates the model's ability to capture the distribution of allele frequency trajectories and illustrates the representational power of deep generative models on the example of linkage disequilibrium (LD) estimation. Inspecting the internally learned representations enables estimating pairwise LD, which is typically inaccessible in Pool-Seq data. Our model provides competitive LD estimation in Pool-Seq data high degree of LD when compared to existing methods.
\end{abstract}




\maketitle

\section{Introduction}
The study of genome-wide variability in natural populations has fascinated scientists since the origins of evolutionary theory. With the advent of advanced genome sequencing technologies over the last century~\cite{seq_technologies}, we now have the capability to analyze genotypes of entire populations over subsequent generations. This wealth of real-world data provides the opportunity to model advanced evolutionary processes. 
At the nucleotide level, this could involve estimating the distribution of future allele frequencies of a population mating under consistent environmental conditions based on empirical observations, or, extracting additional insights from these observations to enhance our understanding of the evolutionary dynamics. In the scope of this work, we focus on the analysis of Evolve and Resequencing (E\&R) experiments that combine sequencing of pooled individuals (Pool-Seq) with replicated experimental evolution. Experimental evolution of model organisms provides a controllable environment reducing the noise from unavailable information like the population size, the environmental pressure variability or unexpected migration events~\cite{experimental_evolution_lenski}. Replicating the same experiment across different populations accounts for stochastic events such as random mating, providing insights into various adaptation strategies. 
Our work focuses on populations with high genetic heterogeneity, where the response to selection pressure is observed through changes in allele frequencies of the founder population~\cite{e_and_r}. Pool-Seq offers a cost- and time-efficient approach to estimate the allele frequencies of a population by sequencing the DNA of a pool of individuals but introduces challenges such as the unavailable linkage disequilibrium (LD) information and additional sampling noise.


A foundational concept in population genetics is the Wright-Fisher (WF) model~\cite{WF_fisher,WF_wright}, a univariate statistical model that describes genetic variation dynamics in a finite, randomly mating population. The model provides a mathematical abstraction as discrete-time finite-space Markov chain to investigate the effects of genetic drift.  
Various evolutionary forces like mutations, migration or selection can be incorporated in the general assumptions of the WF model. The most prominent extensions and the strategies to approximate allele frequency distributions based on the respective model assumptions are summarized and reviewed by Tataru et al.~\cite{WF_distribution}. 
Although the WF model, along with its possible extensions, is valued for its simplicity and interpretability, it relies on simplified assumptions, such as positional independence and the uncertainty associated with parameters estimated from noisy empirical data. 

Deep generative models offer an alternative perspective by learning internal feature representations from multivariate data with minimal prior knowledge. 
However, they have so far received only limited attention in the field of population genomics due to the need for large amounts of training instances and challenges in model interpretability. Generative models like the variational autoencoder (VAE)~\cite{Kingma2014} have so far mainly been applied for the prediction of mutation effects~\cite{DGM_mutations,DGM_disease_variant_pred} or the identification of population clusters~\cite{10.1093/g3journal/jkac020,Meisner_2022}. Battey et al.~\cite{10.1093/g3journal/jkaa036} demonstrated for individual sequencing data in human genomes that VAE-generated latent genotype embeddings effectively capture subtle population structures, while generative adversarial networks (GANs) and restricted Boltzmann machines have been applied to generate artificial genomes that preserve complex genetic features~\cite{GNN_artificial_human,conv_wasserstein_gan}. Specifically, Booker et al.~\cite{Booker2022.09.17.508145} employed a CNN Wasserstein GAN to simulate population genetic alignments, capturing key patterns such as site frequency spectrum, population differences, and LD.

We propose a VAE inspired neural network, which differs from the previous methods in three key aspects. First, we examine the cost-effective Pool-Seq approach instead of Individual-Seq. Pool-Seq aggregates the information of single individuals and consequently entails a higher information loss, such as the loss of LD, resulting in challenging information extraction. Second, we incorporate entire time series of pooled data from replicated evolving populations into our analysis, rather than considering only single time points.
This allows to model the {\em allele frequency distribution trajectory (AFDT)} of future, unseen generations and infer LD information from the internally learned model representations. 
Third, unlike conventional workflows that separate training on simulated data from evaluation on empirical data, our approach leverages a temporally structured dataset, enabling training on initial generations and prediction on subsequent generations within the same population.

Our contributions are the following:
\begin{itemize}
    \item Developing a deep generative model to predict the distribution of AFDTs.
    \item Extracting pairwise LD estimations from the internal network representations.
    \item Analyzing the model performance on dataset with different degree of LD.
    
\end{itemize}
The source code of our method and experimental setup is available at \href{https://github.com/kramerlab/DGMs-of-evolution}{https://github.com/kramerlab/DGMs-of-evolution}.

\section{Methods}

\subsection{E\&R mathematical notation}
To define a generative model for E\&R experiments, we introduce a mathematical framework that formalizes the considered experimental study setup. Consider a population of $N$ (diploid) individuals sexually reproducing over $G$ non-overlapping generations under a constant selection pressure.
The experiment is repeated $R$ times with the same ancestral population. 
Each population is sequenced with Pool-Seq at intervals of $c$ generations for each replicate and aligned to estimated the allele frequencies of the respective population. 
The complete time series dataset $f$ consists of $R\times\frac{G}{c}$ records, with $f_{i,r}^{(g)}$ representing the allele frequency estimates for generation $g$, SNP number $i$ (sorted along the chromosomes) and replicate $r$. 
In our study, we focus on bi-allelic SNPs, as they are the most common and well-characterized variant type in population genetics. 
For each SNP position $i$ we examine the frequency $f_{i,r}^{(g)}\in [0,1]$ of the nucleotide with the highest frequency in the ancestral population. 
To simulate replicated experimental evolution, our goal is to estimate a probability distribution:
\begin{equation*}
h_{c}(f^{(g+c*b))}_{i,r}\mid f^{(g:g+c*b)}_{i,r})
\end{equation*} 
from which the frequencies of the generation step $g+c*b$, conditioned on the time batch $b$ of previous frequencies $f^{(g:g+c*b)}_{i,r}=(f^{(g)}_{i,r},..,f^{(g+c*(b-1))}_{i,r})$, can be sampled.

While the evolutionary process already introduces natural stochastic events like random mating and recombination, Pool-Seq induces additional sequencing noise. 
In the scope of this work, we focus on Pool-Seq noise of two subsequent steps that model the sampling of individuals and alignment coverage as suggested by Jónás et al.~\cite{10.1534/genetics.116.191197}.
Frequency estimates with additional Pool-Seq noise are defined as $\overline{f}_{r,i}^{(g)}$.

\subsection{Linkage disequilibrium}
\begin{figure}
\begin{center}
\includegraphics[width=0.47\textwidth]{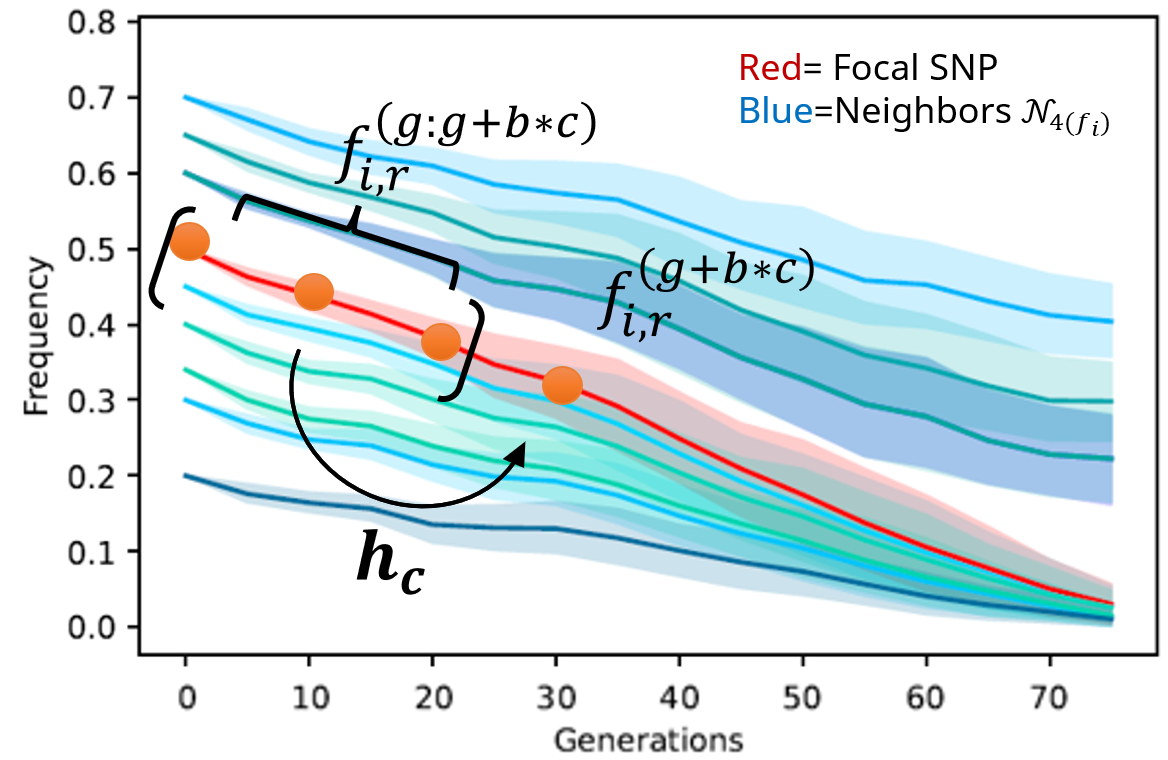}
\caption{AFDTs under negative selection in LD. The red trajectory distribution represents the focal SNP, for which $h_c$ predicts the frequency distribution of the next generation (g+b*c). This prediction can be based on information from both previous generations of the focal SNP and neighboring loci in LD (blue).
}
\label{linkage_demonstration}
\end{center}
\end{figure}

LD refers to the non-random association of two alleles at different loci on a chromosome. 
These associations can result in specific patterns during evolutionary adaptation, such as hitchhiking. 
In the scope of our work, we primarily focus $r^2$~\cite{hill1968linkage}, which is among the most widely used measures for assessing LD between biallelic loci. Let $p_A$ and $p_B$ represent the frequencies of alleles A and B at two distinct loci, respectively, and $p_{AB}$ denote the frequency of the allele pair $(A, B)$. The LD measure $r^2$ is defined as:
\begin{align}
r^2 = \frac{(p_{AB} - p_A p_B)^2}{p_A (1 - p_A) p_B (1 - p_B)}.
\end{align}
As the joint frequency $p_{AB}$ of two loci is not available within Pool-Seq data, direct LD calculation is impossible. 
By observing allele frequencies over subsequent generations, loci in LD will tend to evolve equally (or inversely) to each other offering a pattern with the potential for indirect LD recognition.
While our primary objective is the estimation of future AFDT in the noisy Pool-Seq environment, we hypothesize that incorporating information from neighboring positions (in LD) of a specific SNP, called the focal SNP, can result in learning LD associations to correct the information of the focal SNP's trajectory corrupted by sampling noise. 
We define the focal SNP's $f_{i}$ neighbors, 
falling into the $w$ nearest neighbor region, as $\mathcal{N}_w(f_i) = \left\{f_j \mid j \in \left\{0,1, \dots, n-1\right\}, \, \mid i-j\mid \leq \mid i-w \mid \right\}$. 
Figure~\ref{linkage_demonstration} illustrates the main idea of LD integration in an idealized simulation of a focal SNP $f_i$ (red) and the corresponding neighboring SNPs $\mathcal{N}_4(f_i)$ in LD with the focal SNP. While the focal SNP evolves under negative selection, the AFDT patterns of the neighboring SNPs evolve similarly over time although they are no direct targets of selection. 
AFDT information of neighboring loci could support the prediction of the probability distribution $h_c$, if their LD coefficient with the focal SNP is increased. In the case of additional Pool-Seq sampling noise, the consideration of multivariate relations should support the estimation of univariate SNP trajectories.

To construct a generative deep learning model for AFDT prediction, two sources of information can be considered: the univariate information of the focal SNP over time, which provides the primary trajectory information, and the trajectories of neighboring SNPs, supporting the prediction. Merging the information of those two input sources could reveal which loci are in LD by solving the task of AFDT forward predictions.

\subsection{Generative network architecture}
\begin{figure*}
\begin{center}
\resizebox{0.8\textwidth}{!}{
\input{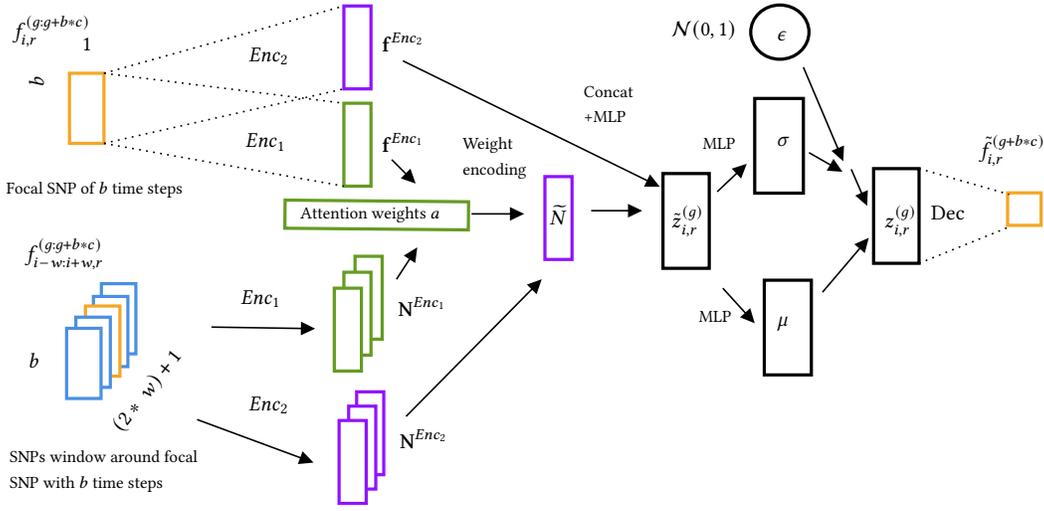}
}
\caption{ Architecture of the VAE-like model. Focal SNP and neighboring loci trajectories are processed by two encoders. $Enc_1$ generates embeddings for attention weights between focal SNP and neighbors to weight the neighbors' contribution for meaningful $z^{(g)}_{i,r}$ embeddings. The decoder predicts the focal SNP frequency of the next generation.}

\label{bi_allelic_architecture}
\end{center}
\end{figure*}

Generative neural networks constitute a powerful class of deep learning models to estimate the underlying distribution of a unknown data generation process. They enable the generation of novel samples that share similar characteristics. 
A well-known architecture within this class is the VAE~\cite{Kingma2014} , which also has the potential to be extended to time series data. Unlike traditional autoencoders, which encode data into a deterministic latent space and decode it to the original input, VAEs introduce a probabilistic approach. They encode the input into a distribution over the latent space, learning the parameters of a prior distribution. During training, a sample is drawn from the learned distribution and decoded to reconstruct the input. The objective of the VAE is to maximizes the likelihood of input reconstruction, while regularizing the learned latent space so that it conforms to the chosen prior distribution. To adapt the VAE structure to time series forecasting, the decoder is modified to predict the allele frequency $\tilde{f}_{i}^{(g+b*c)}$ of the subsequent time step rather than learning an identity map of $\tilde{f}_{i}^{(g:g+b*c)}$. This output can subsequently be processed by the network to forecast the next time step. An overview of our generative network can be found in Figure~\ref{bi_allelic_architecture}.

Contrary to prior research, our approach does not process the entire genome or chromosome in a single training iteration. Instead, we divide the data into smaller, self-contained units, which significantly increases the number of training samples and mitigates memory constraints. By distributing the learning task across numerous smaller, modular instances, this strategy effectively addresses the challenge of small, high-dimensional datasets. Consequently, our model can be trained exclusively on real-world data without relying on potentially inaccurate simulations. Additionally, this approach allows the VAE to directly categorize individual SNP positions without the need to en- and decode uncorrelated information of neighboring SNPs like their starting frequencies.

We propose that the VAE processes not only the focal SNP information, but also a window of fixed length containing the focal SNP in the center. This approach enables leveraging multivariate patterns like LD. 
The pairwise VAE-like model is designed as follows: 
The encoder estimates the posterior distribution $q_{\phi}(z^{(g)}_{i,r}\mid x^{(g)}_{i,r})$ 
by processing the pairwise input $x^{(g)}_{i,r}=(f_{i,r}^{(g:g+b)},f_{i-w:i+w,r}^{(g:g+b)})$. The separation of those two input sources enables processing the trajectory of the focal SNP in relation to the trajectories of its neighbors.
We design the encoder the architecture with an attention based mechanism that aims to measure the similarity between the considered trajectories. Therefore, the encoder consists of two 
subnetworks $Enc_1$ and $Enc_2$. While $Enc_1$ aims to encode the trajectory of a SNP so that the resulting embeddings can be compared with each other with respect to the degree of similarity, $Enc_2$ aims to map the trajectories of the individual SNPs to an embedding space that contains the necessary information to predict the frequency of the next time step. Encoding results to embeddings with $M$ the size of the latent space:
\begin{equation*}
    Enc_1(f_{i,r}^{(g:g+b)})=\mathbf{f}^{Enc_1}\in \mathbb{R}^{1\times M}
\end{equation*}
\begin{equation*}
    Enc_1(f_{i-w:i+w,r}^{(g:g+b)})=\mathbf{N}^{Enc_1}\in \mathbb{R}^{(2w+1)\times M}\text{, }
\end{equation*}
\begin{equation*}
    Enc_2(f_{i,r}^{(g:g+b)})=\mathbf{f}^{Enc_2}\in \mathbb{R}^{1\times M}
\end{equation*}
\begin{equation*}
    Enc_2(f_{i-w:i+w,r}^{(g:g+b)})=\mathbf{N}^{Enc_2}\in \mathbb{R}^{(2w+1)\times M}\text{.}
\end{equation*}
Similarity scores $\mathbf{s}$ are calculated based on $\mathbf{f}^{Enc_1}$ and $\mathbf{N}^{Enc_1}$:
\begin{equation*}
    \mathbf{s} = \frac{\mathbf{f}^{Enc_1}}{\|\mathbf{f}^{Enc_1}\|_2} \cdot \left(\frac{\mathbf{N}^{Enc_1}}{\|\mathbf{N}^{Enc_1}\|_2}\right)^T \in \mathbb{R}^{2w+1}
\end{equation*}
that generate attention weights $\mathbf{a}\in \mathbb{R}^{2w+1}$ with:
\begin{equation*}
    a_i= \frac{\exp(s_i)}{\sum_{j=0}^{2w} \exp(s_j)}.
\end{equation*}
The $\mathbf{N}^{Enc_2}$ is subsequently weighted according to $\mathbf{a}$ to receive $\widetilde{N}=\sum_{i=0}^{2w}a_i \mathbf{N}^{Enc_2}_i $, the encoded neighboring information in relation to the focal SNP.

To evaluate the impact of incorporating neighboring SNP information, we compare two encoder architectures.
The first, termed ``no w'', considers only the focal SNP information, with $\widetilde{z}_{i,r}^{(g)}=\mathbf{f}^{Enc_2}$. 
The second model, termed ``w'', derives $\widetilde{z}_{i,r}^{(g)}$ through a linear combination of $\widetilde{N}$ and $\mathbf{f}^{Enc_2}$ 
by combining focal and neighboring SNP information.

The decoder estimates the next frequency $p_{\theta}(f_{i,r}^{(g+b*c)}\mid z^{(g)}_{i,r})$ by mapping the sampled $z_{i,r}^{(g)}$, through multiple layers of multilayer perceptrons (MLPs) followed by a final sigmoid activation. 
We adapt the expectation lower bound loss (ELBO) loss of the $\beta$-VAE~\cite{higgins2017betavae}:
\begin{align*}
    \mathcal{L}_{1}(x^{(g)}_{i,r},f_{i,r}^{(g+b*c)},\theta,\phi) =\ & \mathbb{E}_{q_{\phi}(z^{(g)}_{i,r}\mid x^{(g)}_{i,r})}\left[\log p_{\theta}(f_{i,r}^{(g+b*c)}|z^{(g)}_{i,r}) \right] \nonumber \\
    &-\beta D_{KL}\left(q_{\phi}(z^{(g)}_{i,r}|x^{(g)}_{i,r})\mid \mid p(z^{(g)}_{i,r})\right)
\end{align*}
with $g\in \left\{0,c,..,c*(n_{max}-1)\right\}$ as objective to maximize.
The reconstruction term is estimated with the $L_{2}$ loss between the expected and the predicted next time step. The Kullback–Leibler divergence (KLD) that regularizes the latent space is scaled with $\beta$. The standard normal distribution is taken as prior $p(z^{(g)}_{i,r})$. Note that by changing the reconstruction task to forecasting, the ELBO can not be guaranteed.

\section{Evaluation setting}
\subsection{Model configurations}
The following hyperparameters are chosen for our deep generative network, based on preliminary experiments: $tanh$ as activation function, a $D_{KL}$ regularizer $\beta$ of 0.0001, window size $w$ of 50, batchsize of 100, a latent dimension $M$ of 10 and ADAM as optimizer. For the encoder architecture, the focal SNP encoding with 3 MLP layers (with dimension 100, 50 and 50) is combined with the neighboring SNP window encoding, which consists of 6 MLP layers (with dimension 100, 50, 25 and 12). The decoder architecture was composed of 3 MLP layers (with dimensions 20, 10 and 5). The network is trained with a learning rate of 0.0001 for 8000 sessions on all SNPs, followed by additional fine-tuning for 8000 epochs with a learning rate of 0.00001 with those SPNs that reach the highest AFC on the training generations (top 10\%). Our model was trained on the Pool-Seq generations 0, 5, 10, ..., 30 including Pool-Seq noise $\overline{f}$ with a time batch size of 6 and evaluated on ground truth simulations $f$ from generations 35, 40, ..., 75.

We compare the results of our deep generative model to an extension of the WF model which includes the simulation of selection. In the later sections, we will refer to this model simply as WF model, even though we are aware that this term is generally only used for neutral evolution. Under the assumption of selection, genotypes are transmitted to the subsequent generation with probabilities determined by their fitness. 
The WF model, describes a finite population undergoing random mating in discrete, non-overlapping generations. To model random mating, allele frequencies are sampled from a Binomial distribution in the bi-allelic case and normalized by $2N_e$ for diploid organisms~\cite{WF_fisher,WF_wright,19710105376,WF_distribution}:
\begin{equation*}
f^{(g+1))}_{i,r}\sim\frac{Bin(2N_e,g(f^{(g))}_{i,r}))}{2N_e}
\end{equation*}
The function $g:[0,1]\rightarrow[0,1]$ extends the general WF model to account for additional evolutionary forces—in this case, selection. The identity map $g(x)=x$ corresponds to the standard WF model, representing genetic drift in the absence of selection.
In our framework, $g(x)$ represents the fitness effects of selection acting on the current allele frequency $f^{(g))}_{i,r}$. The fitness function~\cite{19710105376} is defined as:
\begin{equation*}
g(x)=\frac{(1+s)x^2+(1+\frac{1}{2}s)x(1-x)}{(1+s)x^2+(1+\frac{1}{2}s)x(1-x)+(1-x)^2}
\end{equation*}
where $s$ denotes the selection coefficient. In the case of neutral evolution ($s=0$), the fitness function reduces to the identity map.

Accurate effective population size $N_{e}$ and the selection coefficient $s$ estimation is critical for the quality of the baseline model, as these parameters govern the variance and direction of the AFDT, respectively. While $N_{e}$ is computed as the the average over all replicates with the R package PoolSeq~\cite{10.1534/genetics.116.191197}, $s$ is estimated for every SNP with a linear least-squares regression~\cite{10.1093/molbev/msx225}. $N_{e}$ and $s$ are estimated based on generations 0 and 30.

\subsection{Simulated E\&R description}
The analysis and evaluation of our presented model, requires training data derived from a controllable experimental environment, to understand the underlying evolutionary process and the related results. 
Therefore, we conduct the evaluation on E\&R simulations generated with Mimicree2~\cite{Mimicree2}. The tool performs individual based forward simulations in a discrete time model, enabling the incorporation and monitoring of haplotypes, recombination maps, and target SNPs. 

We employed the quantitative trait model, selecting a set of target SNPs $t_k \in T $ with effect sizes $e_{k}=\frac{2*(k+1)}{\mid T \mid}$ to simulate selection by truncating phenotypic extreme individuals. The evolution of a \textit{Drosophila melanogaster} population on chromosome L2 is simulated with a census size of 1000 individuals over 75 generations, with 10 replicates. After each generation, 99\% of the individuals with the highest pronounced phenotypic value randomly mate for the next generation.

\begin{figure*}
\includegraphics[width=0.98\textwidth]{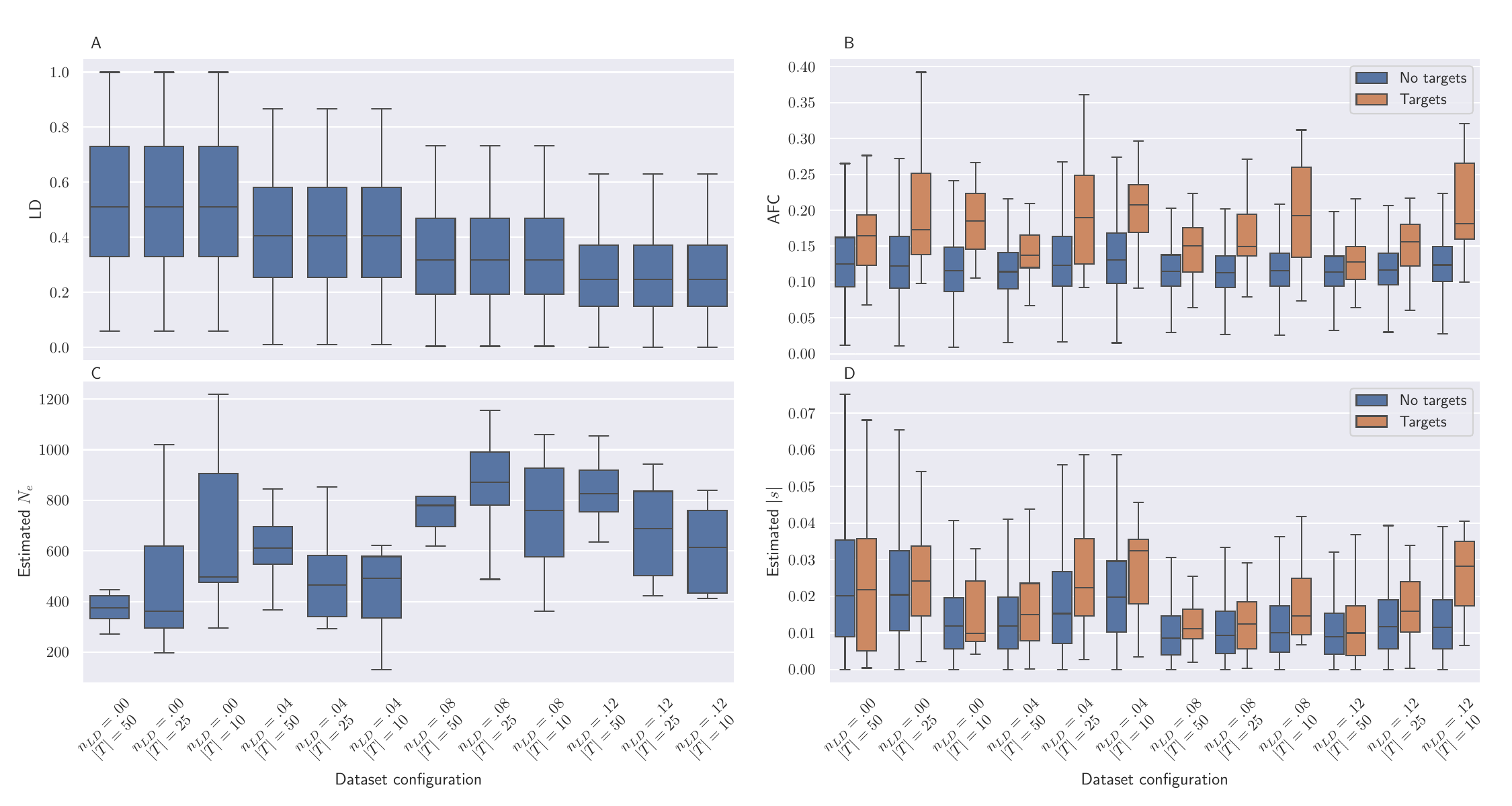}

\caption{Main characteristics of the generated datasets. (A) $r^2$ distributions of 1,000 randomly selected SNP pairs in $\mathcal{N}_{50}$ show decreasing LD with increasing $n_{LD}$. (B) Mean AFC for ``Targets'' and ``No Targets'' across generations 0 and 30, with higher AFC for ``Targets'' and a reduction at higher $n_{LD}$. (C) Estimated effective population size $N_e$ across 10 replicates, increasing with $n_{LD}$. (D) Distribution of estimated absolute values $s$, with high signals for ``No Targets''  compared to ``Targets'' especially visible for datasets with larger number of targets and high LD degree.}
\label{fig:dataset_exp2_charcteristics}
\end{figure*}

To analyze and backtrack the LD consideration of the generative network, we control the frequencies of the ancestral population $f_{i}^{(0)}$, vary the number of target SNPs with $\mid T \mid \in\left\{10,25,50\right\}$, and simulate different degrees of pairwise LD during the generation of artificial haplotypes. 
The $f_{i}^{(0)}$ were sampled from the uniform distribution $ \mathcal{U}(0.05,0.95)$ to avoid values near 0 or 1, preventing rapid SNP fixation. 
The choice of the target positions was driven by the requirement to capture distinct selection patterns. The target SNPs were spaced sufficiently apart to ensure that linked loci were not subject to confounding effects from different selection targets. Consequently, the chromosome was divided into $\mid T \mid$ distinct, equally-sized regions, each containing one randomly selected target SNP with a starting frequency between 0.45 and 0.55. 

The overall degree of LD was controlled by introducing noise into a primary haplotype file characterized by maximal possible LD. Each haplotype file $h\in \left\{0,1\right\}^{\mid C \mid \times2N}$ encodes the binary nucleotide indices for $N$ individuals in a diploid population, across loci $i\in C$. In the primary maximal haplotype file $h_{max}$, the nucleotides for each individual were assigned based on the sampled frequency $f_{i}^{(0)}$. 
Specifically, for each locus $i$, the nucleotide pattern was determined as follows:
\begin{align*}
    \underbrace{0, 0, \ldots, 0,}_{2N f_i^{(0)}} 
\underbrace{1, 1, \ldots, 1}_{2N (1 - f_i^{(0)})}.
\end{align*}
This pattern ensures that the LD between all pairs of loci is maximized. To generate haplotype files $h$ with varying degrees of LD, we introduce the noise parameter $n_{LD}\in \left\{.00,\allowbreak .04,\allowbreak .08,\allowbreak .12\right\}$. For each $h$, this parameter defines the fraction $n_{LD}\times \mid h\mid$ of randomly selected nucleotide positions in $h_{max}$ that are flipped. These random flips introduce noise, reducing the overall degree of pairwise LD.

The Pool-Seq data $f$ extracted from the Mimicree2 simulations is inherently free from sequencing-related noise. To simulate this noise and obtain $\overline{f}$, we followed the two-step sampling approach described by Jónás et al.~\cite{10.1534/genetics.116.191197}. First, sequencing was modeled with a hypergeometric distribution by simulating the random sampling of $n_{sampling}$ individuals from a total population of $N=1000$. 
Subsequently, the average sequencing coverage $n_{cov}$ was simulated with a binomial distribution. 
Both parameters, $n_{sampling}$ and $n_{cov}$, determine the intensity of simulated noise, with lower values leading to a higher noise level. For our analysis, we generated datasets for all $\mid T \mid$ and $n_{LD}$ configurations, consistently setting the noise parameters to $n_{sampling}=100$ and $n_{cov}=40$.

Simulation performance is analyzed on two sets of SPNs: the target SNPs $t_k\in T$ (referred to as ``Targets'') and a random sample of 9.000 positions that are not in $\mathcal{N}_{500}(t_k)$ the 500th neighbor range of any target SNP (referred to as ``No targets''). This distinction facilitates the comparison of distributional differences of each SNP group. ``No targets'' typically exhibit less changes in the distribution mean over generations, while the variance grows over time. In contrast, for ``Targets'' SNPs, changes in the distribution mean of the trajectory are of particular interest.

Figure~\ref{fig:dataset_exp2_charcteristics} summarizes the main characteristics of the generated datasets, providing insights into the model's behavior in the subsequent sections. Subfigure A illustrates the overall degree of LD in the generated haplotype files by displaying the $r^2$ distribution of 10,000 randomly selected SNP pairs located within the $\mathcal{N}_{50}$ neighborhood. As expected, $n_{LD}=.0$ shows the highest possible LD consistent with the sampled starting frequencies, while increasing $n_{LD}$ results in progressively lower and more dispersed LD values. Subfigure B presents the mean allele frequency change (AFC) between generations 0 and 30 for both ``Targets'' and ``No Targets''. As anticipated, the AFC for ``Targets'' is consistently higher than for ``No Targets'' across all datasets. Regardless of $n_{LD}$, the mean AFC of ``Targets'' increases as the number of targets decreases, potentially due to conflicting selection signals among multiple targets. With higher $n_{LD}$, the AFC distributions for ``No Targets'' decrease due to reduced LD with ``Targets'', while the median of these distributions remains relatively stable across all datasets. Subfigure C shows the distribution of the effective population size $N_e$ across 10 replicates for each dataset. As $n_{LD}$ increases, the estimated $N_e$ also rises, reflecting the reduced degree of LD. 
Subfigure D depicts the distribution of the estimated absolute selection coefficient $s$. The Values of the ``No Targets'' are unexpectedly high in comparison to the values of the ``Targets'', an effect especially visible for large number of targets ($|T|=50$), influenced by the degree of LD and the challenges of accurately estimating selection coefficients under noisy conditions.

\section{Results}
\subsection{AFDT generation}
\begin{figure*}
\begin{center}
\includegraphics[width=0.98\textwidth]{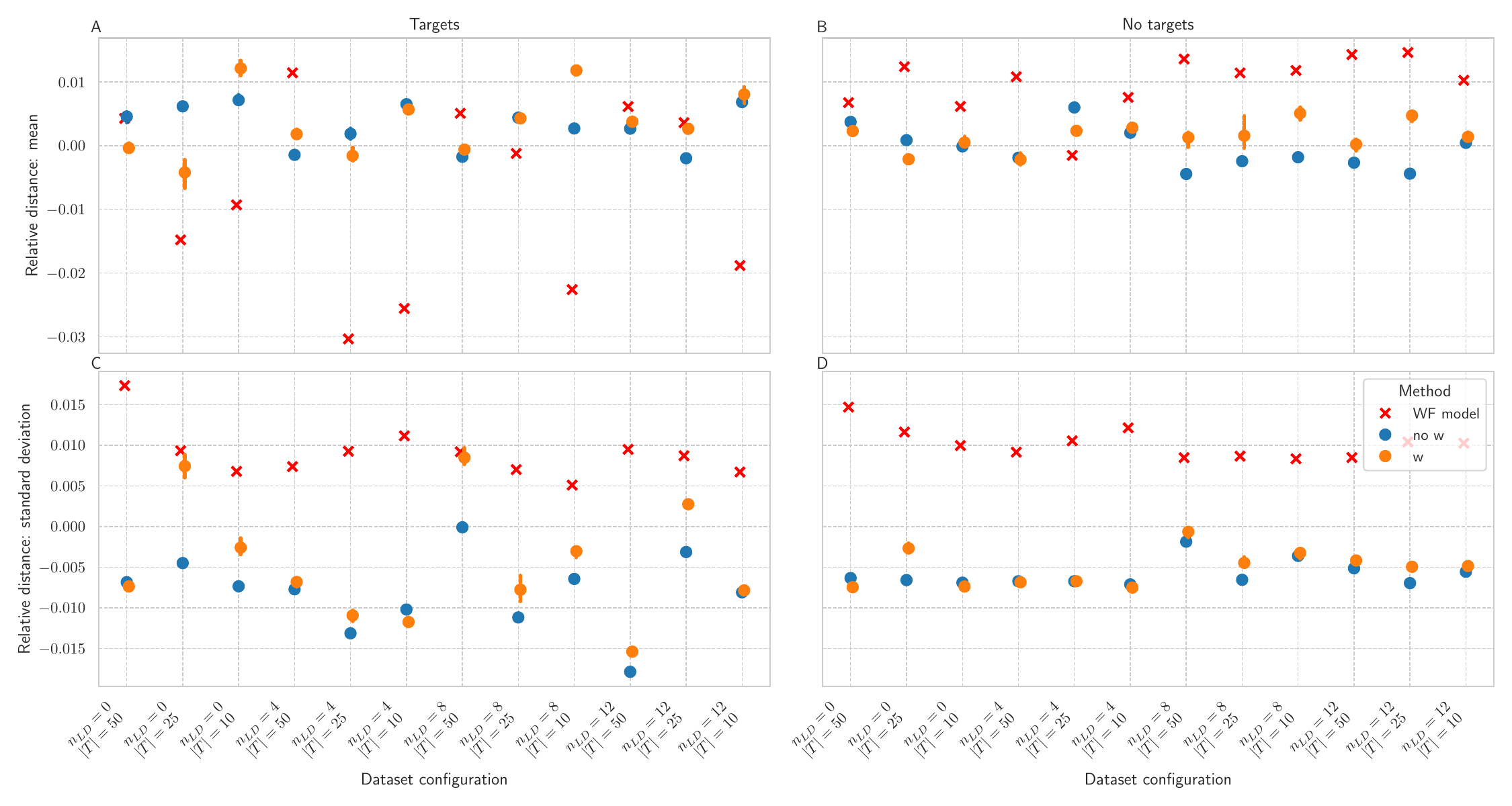}%
\caption{AFDT generation performance across all data configurations. Lower values indicate better performance, with values below 0 outperforming the identity baseline. (A) and (B) show the relative distance of the mean for ``Targets'' and ``No targets'', respectively. VAE models outperform the WF model in three datasets for ``Targets'', match it in two, and underperform in seven. For ``No targets'', VAEs consistently outperform WF. (C) and (D) display the relative distance of the standard deviation. Both VAE models outperform WF, benefiting from more conservative predictions. Including multivariate dependencies (``w'') does not generally enhance performance.}
\label{fig:basic_stats}
\end{center}
\end{figure*}
The performance of the AFDT generation is evaluated by comparing the distribution characteristics of the predictions with the ground truth, focusing on the distribution differences in the mean and standard deviation. In previous experiments, we observed that in environments with minimal changes over the generations, deep neural network predictions can degrade into identity mappings. This degeneration is difficult to detect through simple comparisons, as the lack of variation hides the problem. To address this issue, we introduce the \textit{relative distribution distance}, that assesses how well a model simulates the AFDT for a test generation $g_t+c*j$ compared to a baseline hypothesis. This baseline assumes an identity map, where the AFDT for all test generations is identical to $\overline{f}^{(g_{t})}_{i,r}$, the final training generation $g_t$. The relative distribution distance measures the deviation of the model’s predictions from this baseline, indicating whether the model outperforms or underperforms in comparison to the baseline. It is defined as: 
\begin{align*}
    d_{a_r,j} =\ & \mathbb{E}_i\left[\left| a_{r}\left(f^{(g_t+c\cdot j)}_{i,r}\right)-a_{r}\left(\tilde{f}^{(g_t+c\cdot j)}_{i,r}\right)\right| \right. \nonumber \\
    & \left. -\left| a_{r}\left(f^{(g_t+c\cdot j)}_{i,r}\right)-a_{r}\left(\overline{f}^{(g_{t})}_{i,r}\right)\right| \right]
\end{align*}
with $j>0$ the test generation index. The function $a_{r}$, applied over the replicate index $r$, can represent any aggregation function characterizing a distribution. In our case $a_{r}$ is represented by the mean and the standard deviation. 
The relative distribution distance compares the simulation quality in two terms: while $\mid a_{r}(f^{(g_t+c*j)}_{i,r})-a_{r}(\tilde{f}^{(g_t+c*j)}_{i,r})\mid $ measures the absolute distance from the aggregated ground truth to the aggregated model prediction, the second term $\mid a_{r}(f^{(g_t+c*j)}_{i,r})-a_{r}(\overline{f}^{(g_{t})}_{i,r})\mid $ measures the absolute distance from the aggregated ground truth with the aggregated baseline. 
Negative values of $d_{a_r,j}$ indicate that the model improves upon the baseline, while positive values suggests that the model performs worse. 

Figure~\ref{fig:basic_stats} displays the average relative distance results, including 95\% confidence intervals, on all simulated datasets to evaluate the AFDT generation performance. The results of our VAE configurations ``no w'' and ``w'' are averaged over three independent model runs to account for stochastic effects from random parameter initialization and sampling during training. As the WF model does not rely on random initialization, the result of a single execution are presented.

The relative distance of the mean (A and B) measures how accurately each model tracks changes in allele trajectories across test generations. For ``Targets'' (A), VAE models outperform WF in three datasets, match its performance in two, and underperform in seven. WF excels in datasets with fewer targets and lower LD, where strong AFC signals allow reliable estimation of selection coefficients. However, for ``No Targets'' (B), WF consistently underperforms due to overestimated selection coefficients, an issue likely amplified by the noise. 

Comparing the VAE variants, including neighboring SNP information (``w'') improves performance in only three data\-sets, including two with maximum LD. In most other cases, especially in lower LD datasets (from $n_{LD} = 0.08$), the simpler univariate ``no w'' VAE performs slightly better. This indicates that the noise which affects not only the focal SNP but also its neighbors, makes the added information detrimental to AFDT generation performance in our proposed settings.

The relative distance of the standard deviation captures how well the models approximate the spread of the trajectories within each test generation. 
Especially in the case of smaller $N_e$ estimates, caused by both the increased LD and the induced noise, the WF AFDT's standard deviation is larger than the ground truth AFDTs. 
The VAE models tend to produce more conservative predictions. This enables them to estimate the standard deviation of the trajectory distributions reliably, both for ``Targets'' and ``No targets''. While the choice of VAE model has no effect on the ``No targets'', the univariate ``no w'' variant performs better for `Targets''.

\subsection{LD extraction}
\begin{figure*}
\begin{center}
\includegraphics[width=0.98\textwidth]{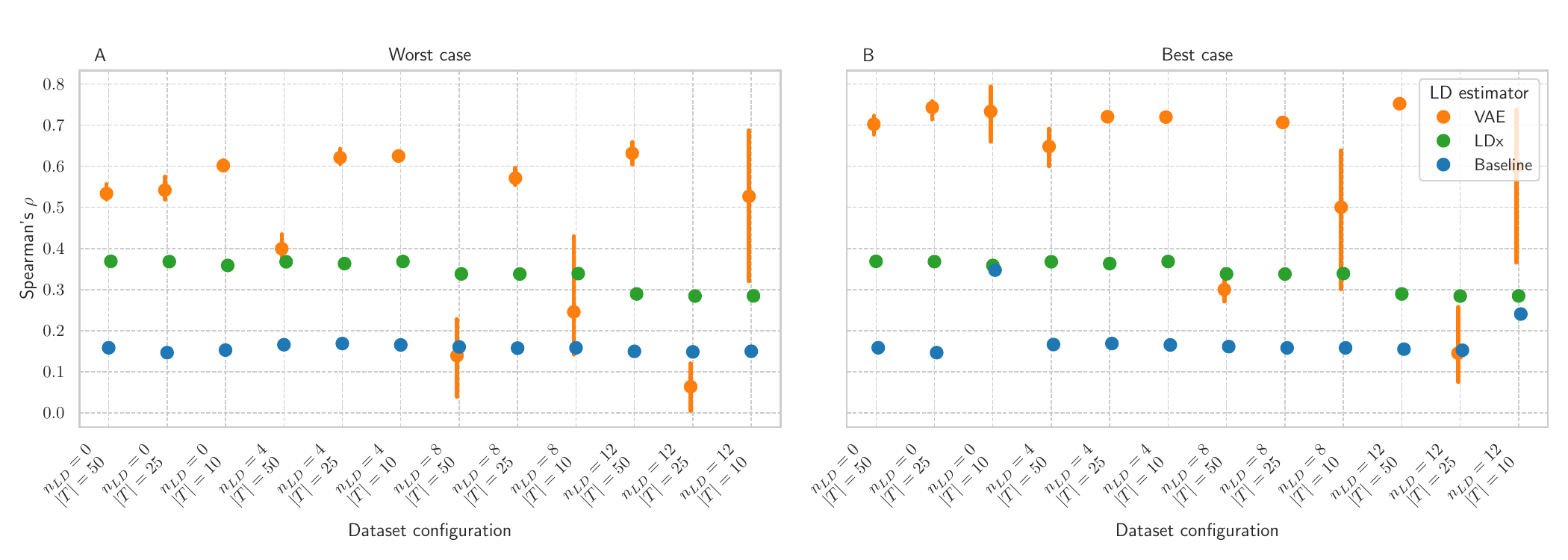}
\caption{LD estimation evaluation with Spearman correlation between estimated and ground truth LD values. (A) shows the worst-case scenario, with correlations calculated across all test SNP pairs. (B) displays the best-case scenario, focusing on SNP pairs with large AFC. Comparing VAE and baseline demonstrates that our model learns representations suitable for LD estimation. Especially on dataset with high LD degree, the VAE outperforms LDx. In datasets characterized by lower LD, the VAE's performance shows increased variability.}
\label{fig:info_of_interest_correlation_example}
\end{center}
\end{figure*}
 
In this section, we evaluate the ability of our proposed deep generative model to extract additional information out of the data. By the proposed semi-transparent VAE ``w'' structure, which calculates similarity scores between the trajectories of the focal SNP and its neighbors, we could investigate this task for the problem of LD estimation. Specifically, similarity scores $\mathbf{s}$ are determined by an attention-based mechanism, where the dot product between the internally learned representations effectively quantifies the directional similarity in the embedding space. 

Our evaluation aims to determine to which extent these similarity scores could infer pairwise LD associations. We compared the VAE’s performance with a reimplementation of LDx~\cite{10.1371/journal.pone.0048588}, 
an LD estimation method for Pool-Seq data and a baseline method. The LDx method in its original form computes the maximally likely $r^2$ conditional on reads that cover both loci.
Since we did not simulate the alignment of individual sequence reads,  and LDx is only applicable to sufficiently long reads, we extended LDx computation to the observed allele frequencies without considering read information. Notably, this modification may reduce LDx performance, but it also enables LD estimation for loci that were previously inaccessible. The baseline computes the absolute value of the original trajectories' scalar product for LD estimation to directly demonstrate the benefit of the model's learned embedding space.

Evaluation was conducted on all test SNPs paired with their $\mathcal{N}_{50}$ neighboring SNPs, resulting in a total of $(\mid T\mid +9000) \times 100$ SNP pairs. For each of the three methods, the mean LD estimation across 10 replicates was chosen as final LD estimate. The ground truth LDs were derived from the ancestral haplotype population. Spearman's $\rho$ correlation between LD estimates and ground truth was calculated to evaluate LD estimation performance. 
The results in Figure~\ref{fig:info_of_interest_correlation_example} present two scenarios: a worst-case scenario (A) considering all SNP pairs and a best-case scenario (B) with the maximal reachable Spearman $\rho$ on a subset of SNP pairs, filtered based on a simple criterion. We assume that SNPs with minor trajectory changes are probably solely influenced by genetic drift, rendering them less informative for robust LD association inference. Specifically, SNP pairs were excluded if they meet the condition:\begin{align*}
\min_{k\in\left\{i,j\right\}}\left(\frac{1}{R}\sum_{r}\mid f_{k,r}^{(0)}-f_{k,r}^{(g)}\mid \right)<\alpha_{AFC},
\end{align*}
for an AFC threshold $\alpha_{AFC}$ .

The results in A demonstrate that the similarity scores derived from the learned trajectory embeddings clearly outperform the baseline in the majority of datasets. This outcome indicates during the VAE’s training process, designed for AFDT simulation, enables it to effectively leverage neighboring SNP information, learning trajectory embeddings optimized for LD estimation. Notably, the VAE outperforms even the LDx method in most dataset configurations. However, its performance exhibits greater variability in datasets with lower LD degree, indicating challenges in maintaining stability in these setting. Furthermore, in contrast to the other methods, the VAE benefits the most from our pre-filtering strategy shown in B, achieving an average increase in Spearman's $\rho$ of 0.146. This procedure enables the advantage of selectively focusing on SNP pairs to provide higher quality LD estimates.

\section{Conclusion}
In this work, the first generative neural network designed for the unsupervised analysis of E\&R experiments has been introduced. The proposed architecture is inspired by VAEs that act on the level of SNPs to predict future AFDTs. We demonstrate that careful model architecture selection enables the possibility of deep learning on E\&R experiments, considered so far unsuitable for deep learning, to model multivariate dependencies in a hypothesis of evolution.

We evaluated two VAE variants: one using only the focal SNP information and another incorporating data from neighboring SNPs. Our experiments on simulated E\&R datasets with varying degrees of LD showed that both models outperformed the WF baseline in AFDT prediction in the majority of the considered cases, particularly in accurately capturing the standard deviation of the trajectories. While the VAE, as a neural network–based approach, exhibits limitations in accurately predicting the trajectory direction of rare target signals, that deviate substantially from the broader SNP distribution and resemble outliers, it offers a flexible and data-driven framework capable of modeling complex patterns. In contrast, the WF model tends to overestimate the trajectory direction of the dominant ``No targets'' class, driven by its overestimation of selection coefficients. While both models face distinct challenges in AFDT prediction, the VAE presents a promising alternative in investigating hidden pattern of evolution.

Although the inclusion of neighboring SNP information did not significantly enhance AFDT prediction performance under our settings, it enabled direct extraction of SNP pair similarity values suitable for LD estimation, information typically unavailable in Pool-Seq data. Our model rivals existing methods for LD estimation in Pool-Seq data, demonstrating the emergence of useful properties as a natural byproduct of the generative modeling approach. 

In future work we would like to investigate to which degree haploblocks can be identified by unsupervised time series comparison. We believe that the potential of internal model representations has not yet been fully exploited. Here we would like to find further latent patterns that have contributed to the network decision. Our first generative deep learning model in the field of E\&R analysis illustrates the potential of such an architecture for future uses.

\begin{acks}
The authors gratefully acknowledge the computing time granted on the supercomputer MOGON II at Johannes Gutenberg University Mainz (hpc.uni-mainz.de). This work was supported by the Research Center For Emergent Algorithmic Intelligence at the University of Mainz funded by the Carl-Zeiss Foundation.

\end{acks}


\begin{thebibliography}{22}


\ifx \showCODEN    \undefined \def \showCODEN     #1{\unskip}     \fi
\ifx \showISBNx    \undefined \def \showISBNx     #1{\unskip}     \fi
\ifx \showISBNxiii \undefined \def \showISBNxiii  #1{\unskip}     \fi
\ifx \showISSN     \undefined \def \showISSN      #1{\unskip}     \fi
\ifx \showLCCN     \undefined \def \showLCCN      #1{\unskip}     \fi
\ifx \shownote     \undefined \def \shownote      #1{#1}          \fi
\ifx \showarticletitle \undefined \def \showarticletitle #1{#1}   \fi
\ifx \showURL      \undefined \def \showURL       {\relax}        \fi
\providecommand\bibfield[2]{#2}
\providecommand\bibinfo[2]{#2}
\providecommand\natexlab[1]{#1}
\providecommand\showeprint[2][]{arXiv:#2}

\bibitem[Ausmees and Nettelblad(2022)]%
        {10.1093/g3journal/jkac020}
\bibfield{author}{\bibinfo{person}{Kristiina Ausmees} {and}
  \bibinfo{person}{Carl Nettelblad}.} \bibinfo{year}{2022}\natexlab{}.
\newblock \showarticletitle{{A deep learning framework for characterization of
  genotype data}}.
\newblock \bibinfo{journal}{\emph{G3 Genes|Genomes|Genetics}}
  \bibinfo{volume}{12}, \bibinfo{number}{3} (\bibinfo{date}{01}
  \bibinfo{year}{2022}).
\newblock
\showISSN{2160-1836}
\href{https://doi.org/10.1093/g3journal/jkac020}{doi:\nolinkurl{10.1093/g3journal/jkac020}}


\bibitem[Battey et~al\mbox{.}(2021)]%
        {10.1093/g3journal/jkaa036}
\bibfield{author}{\bibinfo{person}{C~J Battey}, \bibinfo{person}{Gabrielle~C
  Coffing}, {and} \bibinfo{person}{Andrew~D Kern}.}
  \bibinfo{year}{2021}\natexlab{}.
\newblock \showarticletitle{{Visualizing population structure with variational
  autoencoders}}.
\newblock \bibinfo{journal}{\emph{G3 Genes|Genomes|Genetics}}
  \bibinfo{volume}{11}, \bibinfo{number}{1} (\bibinfo{date}{01}
  \bibinfo{year}{2021}).
\newblock
\showISSN{2160-1836}
\href{https://doi.org/10.1093/g3journal/jkaa036}{doi:\nolinkurl{10.1093/g3journal/jkaa036}}
\newblock
\shownote{jkaa036}.


\bibitem[Booker et~al\mbox{.}(2023)]%
        {Booker2022.09.17.508145}
\bibfield{author}{\bibinfo{person}{William~W Booker}, \bibinfo{person}{Dylan~D
  Ray}, {and} \bibinfo{person}{Daniel~R Schrider}.}
  \bibinfo{year}{2023}\natexlab{}.
\newblock \showarticletitle{{This population does not exist: learning the
  distribution of evolutionary histories with generative adversarial
  networks}}.
\newblock \bibinfo{journal}{\emph{Genetics}} \bibinfo{volume}{224},
  \bibinfo{number}{2} (\bibinfo{date}{04} \bibinfo{year}{2023}).
\newblock
\showISSN{1943-2631}
\href{https://doi.org/10.1093/genetics/iyad063}{doi:\nolinkurl{10.1093/genetics/iyad063}}
\newblock
\shownote{Article ID: iyad063}.


\bibitem[Crow and Kimura(1970)]%
        {19710105376}
\bibfield{author}{\bibinfo{person}{J.~F. Crow} {and} \bibinfo{person}{M.
  Kimura}.} \bibinfo{year}{1970}\natexlab{}.
\newblock \bibinfo{booktitle}{\emph{An introduction to population genetics
  theory.}}
\newblock \bibinfo{publisher}{New York, Evanston and London: Harper \& Row,
  Publishers}. 1--591 pages.
\newblock


\bibitem[Feder et~al\mbox{.}(2012)]%
        {10.1371/journal.pone.0048588}
\bibfield{author}{\bibinfo{person}{Alison~F. Feder}, \bibinfo{person}{Dmitri~A.
  Petrov}, {and} \bibinfo{person}{Alan~O. Bergland}.}
  \bibinfo{year}{2012}\natexlab{}.
\newblock \showarticletitle{LDx: Estimation of Linkage Disequilibrium from
  High-Throughput Pooled Resequencing Data}.
\newblock \bibinfo{journal}{\emph{PLOS ONE}} \bibinfo{volume}{7},
  \bibinfo{number}{11} (\bibinfo{date}{11} \bibinfo{year}{2012}),
  \bibinfo{pages}{1--7}.
\newblock
\href{https://doi.org/10.1371/journal.pone.0048588}{doi:\nolinkurl{10.1371/journal.pone.0048588}}


\bibitem[Fisher(1930)]%
        {WF_fisher}
\bibfield{author}{\bibinfo{person}{R.~A. Fisher}.}
  \bibinfo{year}{1930}\natexlab{}.
\newblock \bibinfo{booktitle}{\emph{The genetical theory of natural
  selection}}.
\newblock \bibinfo{publisher}{Oxford: Clarendon}.
\newblock


\bibitem[Frazer et~al\mbox{.}(2021)]%
        {DGM_disease_variant_pred}
\bibfield{author}{\bibinfo{person}{Jonathan Frazer}, \bibinfo{person}{Pascal
  Notin}, \bibinfo{person}{Mafalda Dias}, \bibinfo{person}{Aidan Gomez},
  \bibinfo{person}{Joseph Min}, \bibinfo{person}{Kelly Brock},
  \bibinfo{person}{Yarin Gal}, {and} \bibinfo{person}{Debora Marks}.}
  \bibinfo{year}{2021}\natexlab{}.
\newblock \showarticletitle{Disease variant prediction with deep generative
  models of evolutionary data}.
\newblock \bibinfo{journal}{\emph{Nature}}  \bibinfo{volume}{599}
  (\bibinfo{date}{11} \bibinfo{year}{2021}).
\newblock
\href{https://doi.org/10.1038/s41586-021-04043-8}{doi:\nolinkurl{10.1038/s41586-021-04043-8}}


\bibitem[Higgins et~al\mbox{.}(2017)]%
        {higgins2017betavae}
\bibfield{author}{\bibinfo{person}{Irina Higgins}, \bibinfo{person}{Loic
  Matthey}, \bibinfo{person}{Arka Pal}, \bibinfo{person}{Christopher Burgess},
  \bibinfo{person}{Xavier Glorot}, \bibinfo{person}{Matthew Botvinick},
  \bibinfo{person}{Shakir Mohamed}, {and} \bibinfo{person}{Alexander
  Lerchner}.} \bibinfo{year}{2017}\natexlab{}.
\newblock \showarticletitle{beta-{VAE}: Learning Basic Visual Concepts with a
  Constrained Variational Framework}. In
  \bibinfo{booktitle}{\emph{International Conference on Learning
  Representations}}.
\newblock
\urldef\tempurl%
\url{https://openreview.net/forum?id=Sy2fzU9gl}
\showURL{%
\tempurl}


\bibitem[Hill and Robertson(1968)]%
        {hill1968linkage}
\bibfield{author}{\bibinfo{person}{WG Hill} {and} \bibinfo{person}{Alan
  Robertson}.} \bibinfo{year}{1968}\natexlab{}.
\newblock \showarticletitle{Linkage disequilibrium in finite populations}.
\newblock \bibinfo{journal}{\emph{Theoretical and applied genetics}}
  \bibinfo{volume}{38} (\bibinfo{year}{1968}), \bibinfo{pages}{226--231}.
\newblock


\bibitem[Jónás et~al\mbox{.}(2016)]%
        {10.1534/genetics.116.191197}
\bibfield{author}{\bibinfo{person}{Agnes Jónás}, \bibinfo{person}{Thomas
  Taus}, \bibinfo{person}{Carolin Kosiol}, \bibinfo{person}{Christian
  Schlötterer}, {and} \bibinfo{person}{Andreas Futschik}.}
  \bibinfo{year}{2016}\natexlab{}.
\newblock \showarticletitle{{Estimating the Effective Population Size from
  Temporal Allele Frequency Changes in Experimental Evolution}}.
\newblock \bibinfo{journal}{\emph{Genetics}} \bibinfo{volume}{204},
  \bibinfo{number}{2} (\bibinfo{date}{10} \bibinfo{year}{2016}),
  \bibinfo{pages}{723--735}.
\newblock
\showISSN{1943-2631}
\href{https://doi.org/10.1534/genetics.116.191197}{doi:\nolinkurl{10.1534/genetics.116.191197}}


\bibitem[Kawecki et~al\mbox{.}(2012)]%
        {experimental_evolution_lenski}
\bibfield{author}{\bibinfo{person}{Tadeusz Kawecki}, \bibinfo{person}{Richard
  Lenski}, \bibinfo{person}{Dieter Ebert}, \bibinfo{person}{Brian Hollis},
  \bibinfo{person}{Isabelle Olivieri}, {and} \bibinfo{person}{Michael
  Whitlock}.} \bibinfo{year}{2012}\natexlab{}.
\newblock \showarticletitle{Experimental Evolution}.
\newblock \bibinfo{journal}{\emph{Trends in ecology \& evolution}}
  \bibinfo{volume}{27} (\bibinfo{date}{07} \bibinfo{year}{2012}),
  \bibinfo{pages}{547--60}.
\newblock
\href{https://doi.org/10.1016/j.tree.2012.06.001}{doi:\nolinkurl{10.1016/j.tree.2012.06.001}}


\bibitem[Kingma and Welling(2014)]%
        {Kingma2014}
\bibfield{author}{\bibinfo{person}{Diederik~P. Kingma} {and}
  \bibinfo{person}{Max Welling}.} \bibinfo{year}{2014}\natexlab{}.
\newblock \showarticletitle{{Auto-Encoding Variational Bayes}}. In
  \bibinfo{booktitle}{\emph{2nd International Conference on Learning
  Representations, {ICLR} 2014, Banff, AB, Canada, April 14-16, 2014,
  Conference Track Proceedings}}.
\newblock
\showeprint[arXiv]{http://arxiv.org/abs/1312.6114v10}


\bibitem[Mardis(2017)]%
        {seq_technologies}
\bibfield{author}{\bibinfo{person}{Elaine~R. Mardis}.}
  \bibinfo{year}{2017}\natexlab{}.
\newblock \showarticletitle{DNA sequencing technologies: 2006-2016}.
\newblock \bibinfo{journal}{\emph{Nature Protocols}} \bibinfo{volume}{12},
  \bibinfo{number}{2} (\bibinfo{date}{02} \bibinfo{year}{2017}),
  \bibinfo{pages}{213--218}.
\newblock
\showISBNx{17542189}
\urldef\tempurl%
\url{https://www.proquest.com/scholarly-journals/dna-sequencing-technologies-2006-2016/docview/1865214414/se-2}
\showURL{%
\tempurl}


\bibitem[Meisner and Albrechtsen(2022)]%
        {Meisner_2022}
\bibfield{author}{\bibinfo{person}{Jonas Meisner} {and} \bibinfo{person}{Anders
  Albrechtsen}.} \bibinfo{year}{2022}\natexlab{}.
\newblock \showarticletitle{Haplotype and population structure inference using
  neural networks in whole-genome sequencing data}.
\newblock \bibinfo{journal}{\emph{Genome research}}  \bibinfo{volume}{32}
  (\bibinfo{date}{07} \bibinfo{year}{2022}).
\newblock
\href{https://doi.org/10.1101/gr.276813.122}{doi:\nolinkurl{10.1101/gr.276813.122}}


\bibitem[Riesselman et~al\mbox{.}(2018)]%
        {DGM_mutations}
\bibfield{author}{\bibinfo{person}{Adam Riesselman}, \bibinfo{person}{John
  Ingraham}, {and} \bibinfo{person}{Debora Marks}.}
  \bibinfo{year}{2018}\natexlab{}.
\newblock \showarticletitle{Deep generative models of genetic variation capture
  the effects of mutations}.
\newblock \bibinfo{journal}{\emph{Nature Methods}}  \bibinfo{volume}{15}
  (\bibinfo{date}{10} \bibinfo{year}{2018}).
\newblock
\href{https://doi.org/10.1038/s41592-018-0138-4}{doi:\nolinkurl{10.1038/s41592-018-0138-4}}


\bibitem[Schl\"otterer et~al\mbox{.}(2014)]%
        {e_and_r}
\bibfield{author}{\bibinfo{person}{Christian Schl\"otterer},
  \bibinfo{person}{Robert Kofler}, \bibinfo{person}{Elisabetta Versace},
  \bibinfo{person}{Ray Tobler}, {and} \bibinfo{person}{Susanne Franssen}.}
  \bibinfo{year}{2014}\natexlab{}.
\newblock \showarticletitle{Combining experimental evolution with
  next-generation sequencing: a powerful tool to study adaptation from standing
  genetic variation}.
\newblock \bibinfo{journal}{\emph{Heredity}}  \bibinfo{volume}{0}
  (\bibinfo{date}{10} \bibinfo{year}{2014}).
\newblock
\href{https://doi.org/10.1038/hdy.2014.86}{doi:\nolinkurl{10.1038/hdy.2014.86}}


\bibitem[Tataru et~al\mbox{.}(2016)]%
        {WF_distribution}
\bibfield{author}{\bibinfo{person}{Paula Tataru}, \bibinfo{person}{Maria
  Simonsen}, \bibinfo{person}{Thomas Bataillon}, {and} \bibinfo{person}{Asger
  Hobolth}.} \bibinfo{year}{2016}\natexlab{}.
\newblock \showarticletitle{{Statistical Inference in the Wright–Fisher Model
  Using Allele Frequency Data}}.
\newblock \bibinfo{journal}{\emph{Systematic Biology}} \bibinfo{volume}{66},
  \bibinfo{number}{1} (\bibinfo{date}{08} \bibinfo{year}{2016}),
  \bibinfo{pages}{e30--e46}.
\newblock
\showISSN{1063-5157}
\href{https://doi.org/10.1093/sysbio/syw056}{doi:\nolinkurl{10.1093/sysbio/syw056}}


\bibitem[Taus et~al\mbox{.}(2017)]%
        {10.1093/molbev/msx225}
\bibfield{author}{\bibinfo{person}{Thomas Taus}, \bibinfo{person}{Andreas
  Futschik}, {and} \bibinfo{person}{Christian Schl\"otterer}.}
  \bibinfo{year}{2017}\natexlab{}.
\newblock \showarticletitle{{Quantifying Selection with Pool-Seq Time Series
  Data}}.
\newblock \bibinfo{journal}{\emph{Molecular Biology and Evolution}}
  \bibinfo{volume}{34}, \bibinfo{number}{11} (\bibinfo{date}{08}
  \bibinfo{year}{2017}), \bibinfo{pages}{3023--3034}.
\newblock
\showISSN{0737-4038}
\href{https://doi.org/10.1093/molbev/msx225}{doi:\nolinkurl{10.1093/molbev/msx225}}


\bibitem[Vlachos and Kofler(2018)]%
        {Mimicree2}
\bibfield{author}{\bibinfo{person}{Christos Vlachos} {and}
  \bibinfo{person}{Robert Kofler}.} \bibinfo{year}{2018}\natexlab{}.
\newblock \showarticletitle{MimicrEE2: Genome-wide forward simulations of
  Evolve and Resequencing studies}.
\newblock \bibinfo{journal}{\emph{PLOS Computational Biology}}
  \bibinfo{volume}{14}, \bibinfo{number}{8} (\bibinfo{date}{08}
  \bibinfo{year}{2018}), \bibinfo{pages}{1--10}.
\newblock
\href{https://doi.org/10.1371/journal.pcbi.1006413}{doi:\nolinkurl{10.1371/journal.pcbi.1006413}}


\bibitem[Wright(1931)]%
        {WF_wright}
\bibfield{author}{\bibinfo{person}{Sewall Wright}.}
  \bibinfo{year}{1931}\natexlab{}.
\newblock \showarticletitle{Evolution in Mendelian Populations}.
\newblock \bibinfo{journal}{\emph{Genetics}} \bibinfo{volume}{16},
  \bibinfo{number}{2} (\bibinfo{date}{03} \bibinfo{year}{1931}),
  \bibinfo{pages}{97--159}.
\newblock
\showISSN{1943-2631}
\href{https://doi.org/10.1093/genetics/16.2.97}{doi:\nolinkurl{10.1093/genetics/16.2.97}}


\bibitem[Yelmen et~al\mbox{.}(2023)]%
        {conv_wasserstein_gan}
\bibfield{author}{\bibinfo{person}{Burak Yelmen}, \bibinfo{person}{Aurelien
  Decelle}, \bibinfo{person}{Leila Boulos}, \bibinfo{person}{Antoine
  Szatkownik}, \bibinfo{person}{Cyril Furtlehner}, \bibinfo{person}{Guillaume
  Charpiat}, {and} \bibinfo{person}{Flora Jay}.}
  \bibinfo{year}{2023}\natexlab{}.
\newblock \showarticletitle{Deep convolutional and conditional neural networks
  for large-scale genomic data generation}.
\newblock \bibinfo{journal}{\emph{PLOS Computational Biology}}
  \bibinfo{volume}{19} (\bibinfo{date}{10} \bibinfo{year}{2023}),
  \bibinfo{pages}{e1011584}.
\newblock
\href{https://doi.org/10.1371/journal.pcbi.1011584}{doi:\nolinkurl{10.1371/journal.pcbi.1011584}}


\bibitem[Yelmen et~al\mbox{.}(2021)]%
        {GNN_artificial_human}
\bibfield{author}{\bibinfo{person}{Burak Yelmen}, \bibinfo{person}{Aurélien
  Decelle}, \bibinfo{person}{Linda Ongaro}, \bibinfo{person}{Davide Marnetto},
  \bibinfo{person}{Corentin Tallec}, \bibinfo{person}{Francesco Montinaro},
  \bibinfo{person}{Cyril Furtlehner}, \bibinfo{person}{Luca Pagani}, {and}
  \bibinfo{person}{Flora Jay}.} \bibinfo{year}{2021}\natexlab{}.
\newblock \showarticletitle{Creating artificial human genomes using generative
  neural networks}.
\newblock \bibinfo{journal}{\emph{PLOS Genetics}} \bibinfo{volume}{17},
  \bibinfo{number}{2} (\bibinfo{date}{02} \bibinfo{year}{2021}),
  \bibinfo{pages}{1--22}.
\newblock
\href{https://doi.org/10.1371/journal.pgen.1009303}{doi:\nolinkurl{10.1371/journal.pgen.1009303}}


\end{thebibliography}

\end{document}